\documentclass[conference]{IEEEtran}
\IEEEoverridecommandlockouts
\usepackage{cite}
\usepackage{amsmath,amssymb,amsfonts}
\usepackage{algorithm}
\usepackage{algorithmic}
\usepackage{graphicx}
\usepackage{subfigure}
\usepackage{textcomp}
\usepackage{xcolor}
\usepackage{subfig}
\DeclareMathOperator{\st}{s.t.}
\DeclareMathOperator{\sign}{sign}

\def\BibTeX{{\rm B\kern-.05em{\sc i\kern-.025em b}\kern-.08em
    T\kern-.1667em\lower.7ex\hbox{E}\kern-.125emX}}

\title{Query-Efficient Black-Box Attack by\\ Active Learning}

\author{\IEEEauthorblockN{Pengcheng Li}
\IEEEauthorblockA{National Key Laboratory\\
for Novel Software Technology\\
Nanjing University, Nanjing 210023, China\\
Email: lipc@lamda.nju.edu.cn}
\and
\IEEEauthorblockN{Jinfeng Yi}
\IEEEauthorblockA{JD AI Research\\
Email: yijinfeng@jd.com}
\and
\IEEEauthorblockN{Lijun Zhang}
\IEEEauthorblockA{National Key Laboratory\\
for Novel Software Technology\\
Nanjing University, Nanjing 210023, China\\
Email: zhanglj@lamda.nju.edu.cn}}

\begin{document}

\maketitle
\begin{abstract}
Deep neural network (DNN) as a popular machine learning model is found to be vulnerable to adversarial attack. 
This attack constructs adversarial examples by adding small perturbations to the raw input, while appearing unmodified to human eyes but will be misclassified by a well-trained classifier. 
In this paper, we focus on the black-box attack setting where attackers have almost no access to the underlying models. 
To conduct black-box attack, a popular approach aims to train a substitute model based on the information queried from the target DNN. 
The substitute model can then be attacked using existing white-box attack approaches, and the generated adversarial examples will be used to attack the target DNN. 
Despite its encouraging results, this approach suffers from poor query efficiency, i.e., attackers usually needs to query a huge amount of times to collect enough information for training an accurate substitute model. 
To this end, we first utilize state-of-the-art white-box attack methods to generate samples for querying, and then introduce an active learning strategy to significantly reduce the number of queries needed. 
Besides, we also propose a diversity criterion to avoid the sampling bias. 
Our extensive experimental results on MNIST and CIFAR-10 show that the proposed method can reduce more than $90\%$ of queries while preserve attacking success rates and obtain an accurate substitute model which is more than $85\%$ similar with the target oracle. 
\end{abstract}

\section{Introduction}
Deep neural networks (DNNs) have achieved great successes in a variety of domains~\cite{lecun2015deep}. 
However, recent studies have shown that DNNs may be easily fooled by adversarial examples~\cite{goodfellow2014explaining}. For example, in the context of image classification, an adversarial example is an image that is visually indistinguishable to the original image but can mislead the DNN model to output incorrect labels. In addition to image classification, attacks to other DNN-related tasks have also been actively investigated, such as semantic segmentation~\cite{metzen2017universal}, machine translation~\cite{cheng2018seq2sick}, visual QA~\cite{xu2017can}, image captioning~\cite{chen2017show}, speech recognition~\cite{carlini2018audio}, medical prediction~\cite{sun2018identify}, and autonomous driving~\cite{DBLP:journals/corr/EvtimovEFKLPRS17}.

\begin{figure}[ht] 

\centering
\includegraphics[width=8.5cm]{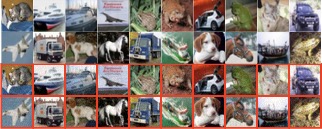}
\caption{Adversarial example generated by our substitute model with FGSM ($\lambda = 0.03$) on CIFAR-10. The first two lines are the original images and the last two lines which are surrounded by red lines are adversarial examples which will mislead the oracle.}
\label{fig:2} 
\end{figure} 

Depending on how much information the attackers have access to, adversarial attack can be broadly classified into two categories: white-box attack and black-box attack. The adversary in the white-box setting has full access to the target DNN model~\cite{goodfellow2014explaining,carlini2017towards,moosavi2016deepfool}. In the black-box setting, adversaries can only access the input and output of the underlying DNN but not its internal configurations and parameters~\cite{chen2017zoo,papernot2017practical}. Recent studies have shown that both of these two categories of attacks can reach a extremely high success rate of attack. Although a lot of defense methods \cite{buckman2018thermometer,guo2017countering,xie2017mitigating} were designed to increase the robustness of the model, the white-box attack \cite{athalye2018obfuscated} can still conquer the model with nearly $100\%$ success rate by estimating the gradient through approximation or expectation \cite{athalye2017synthesizing}.

Compared to the white-box setting, the black-box setting is much more practical since a majority of real-world learning systems do not allow white-box access due to security reasons. Most of existing black-box attack methods are based on the transferability phenomenon~\cite{papernot2017practical}, where an adversary first trains a substitute model and then crafts adversarial examples against it, hoping that the generated adversarial examples can also successfully attack the underlying black-box models. 
Black-box attack can also bypass most defense methods that change the model structure to increase robustness for the reason that it is isolated from the target model. The black-box variants of JSMA \cite{narodytska2016simple} and of the Carlini \& Wagner attack \cite{carlini2017towards} both obtain over $95\%$ success rate on adversarial examples. 
However, a key limitation of these approaches is that  training a substitute network requires a large number of queries to collect sufficient information. For example, the number of queries in~\cite{papernot2017practical} increases almost exponentially with respect to the number of iterations.

In this paper, we address this issue by employing the active learning strategy. Specifically,  
we first utilize the state-of-the-art white-box attack methods to generate adversarial examples. We then improve the query-efficiency of transfer-based framework by actively selecting the most informative samples. Furthermore, we propose a diversity criterion to avoid the bias caused by active learning. We summarize our main contributions as follows:\par
\begin{itemize}
	\item We propose to use more advanced methods for data augmentation in transfer-based framework, and verify that C\&W attack method \cite{carlini2017towards} and Deepfool \cite{moosavi2016deepfool} are more effective than the raw Jacobian-based method  \cite{papernot2017practical}.
	\item We propose to use active learning strategy to select the most informative samples for querying. To avoid the bias caused by active learning, we further introduce a diversity criterion to ensure that the sampled queries are both informative and diverse.
	\item We conduct extensive experiments to evaluate our method. Our empirical results show that the proposed approach can significantly reduce the number of queries while preserve the success rate of attack.\end{itemize}

\section{Related Work} 
In this section, we briefly review the existing work of adversarial attacks and active learning.
\subsection{Adversarial Attack Methods}
\noindent\textbf{White-box attack:}
In the following, we summarize four principal white-box attack methods as follows.

\begin{itemize}
	\item \textbf{Fast Gradient Sign Method (FGSM)} \cite{goodfellow2014explaining}: FGSM aims to construct the adversarial examples such that the $\ell_{\infty}$-norm of the perturbation is constrained by a small constant. The perturbation can be formulated as 
	$$\pmb{\delta}=\lambda \cdot \sign(\nabla_{\mathbf{x}}f(\mathbf{x})),$$ where $\lambda$ is the magnitude of the perturbation and $f(\cdot)$ is the prediction of a neural network. This perturbation can be calculated within one back-propagation of the network. This method affirms that the linear part of the neural network has a serious impact on the composition of the adversarial examples. At the same time, there are no efficient ways to defend against such attacks in neural networks, such as dropout as well as pre-training and adding infinite norm regularizations, other than adding the adversarial examples into the training set.

	\item  \textbf{Jacobian-based Saliency Map Attack (JSMA)} \cite{papernot2016limitations}: JSMA first computes Jacobian matrix of given sample $\mathbf{x}$ as $\nabla_xf(\mathbf{x})$, it then identifies features of $\mathbf{x}$ that made most significant changes to the output. A small perturbation was designed to successfully induce large output variations so that changes in a small portion of features could fool the neural network. For an input $\mathbf{x}$ and a neural network $F(\cdot)$, the output for class j is denoted $F_j (\mathbf{x})$. This attack is constructed by exploiting the adversarial saliency map, which is defined as
$$S(\mathbf{x},t)[i]=
\begin{cases}
0,\text{if}\frac{\partial F_t(\mathbf{x})}{\partial\mathbf{x}_i}< 0 \text{ or } \sum_{j \ne t}\frac{\partial F_j(\mathbf{x})}{\partial\mathbf{x}_i}>0,\\
(\frac{\partial F_t(\mathbf{x})}{\partial\mathbf{x}_i})|\sum_{j \ne t}\frac{\partial F_j(\mathbf{x})}{\partial\mathbf{x}_i}|, \text{otherwise},
\end{cases}
$$
where $t$ is the target class and the procedure of the attack is to iteratively locate the pair of features $\{i, j\}$ that maximizes $S(X, t)[i] + S(X, t)[j]$, and repeatedly perturb the original example by a constant offset.

\item \textbf{DeepFool} \cite{moosavi2016deepfool}: Deepfool assumes that the neural networks are totally linear and there exists a hyperplane which can separate each class from another. The authors derive the optimal solution to this simplified problem and they use a $\ell_2$ minimization-based formulation to search for adversarial examples. The purpose of Deepfool is to solve the following optimization problem:  
	$$
	\min \|\pmb\delta\|_2,\quad \st\quad f(\mathbf{x}+\pmb\delta)\ne f(\mathbf{x})
	$$\par
	DeepFool uses geometry information for directing the search for the minimal perturbation to fool a network.\par
	
	\item \textbf{C\&W attack} \cite{carlini2017towards}: C\&W attack is by far one of the strongest attacks: it can achieve almost $100\%$ attack success rate and has bypassed $10$ different methods designed for detecting adversarial examples~\cite{DBLP:journals/corr/CarliniW17}. It formulates the task of generating adversarial examples as an optimization problem:
	$$  \quad\min \quad  \|\pmb{\delta}\|_p+c \cdot g(\mathbf{x}+\pmb{\delta})
	$$
	$$
	  \st \quad  \mathbf{x}+\pmb{\delta} \in[0,1]^n,\\
	$$
  where $g(\cdot)$ represents an objective function which is designed to simplify the non-linear constraint in the original problem. 
C\&W designed a special objective function based on the hinge loss: $$f(\mathbf{x})=\max\{\max_{i\ne t}[Z(\mathbf{x})]_i-[Z(\mathbf{x})]_t,-k\}.$$ 
C\&W attack also has some variations. For example, EAD attack~\cite{chen2017ead} proposes to use elastic-net regularization, a linear combination of $\ell_1$-norm and $\ell_2$-norm, to replace the $\ell_p$-norm introduced in C\&W attack for penalizing large distortion between the original and adversarial examples.
\par
\end{itemize}

\noindent\textbf{Black-box attack:}
Black-box attack can be roughly divided into the following categories:
\begin{itemize}
	\item \textbf{Score-based attack:} Score-based attack methods rely on the predicted scores (e.g. class probabilities or logits) to estimate the underlying gradients. Typical methods in this category include black-box variants of JSMA
\cite{narodytska2016simple} and of the C\&W attack \cite{chen2017zoo,tu2018autozoom}, as well as generator networks that predict adversarial \cite{hayes2017machine}. 
\item \textbf{Decision-based attack:}  Decision-based attack was proposed by~\cite{brendel2017decision}. It starts from a large adversarial perturbation and then iteratively reduces the perturbation while ensuring that the adversarial example stays in the adversarial region. 
\item \textbf{Transfer-based attack:} Transfer-based attack aims to train a substitute network using the information queried from the underlying target model~\cite{papernot2017practical}. This method uses inputs synthetically crafted by the local substitute model and labeled by the target oracle. 
It has been shown that when the substitute model was well-trained, its adversarial examples can also mislead the target model.  The previous work of transfer-based attack \cite{papernot2017practical} used Jacobian-based data augmentation, so the number of query doubles with the increase of iterations.  Although it used reservoir sampling method which is one of the random select methods to reduce the queries, the exponential growth in the first few iterations has led to a significant increase in the number of queries. 
\end{itemize}

\subsection{Active Learning}
Active learning is a widely used framework in which the learner is able to select the most informative unlabeled examples for human annotation \cite{settles2010active}.
The learning algorithm actively engages an oracle to request information in addition to the original training set. The learner employs a query strategy to select instances for labeling. The main purpose of active learning is to reduce the total cost of labeling. It can be roughly divided into three scenarios: query synthesis, selective sampling and pool-based active learning \cite{zhu2008active,small2010margin,har2003constraint}.  Pool-based active learning is a practical scenario, where learners can choose from the pool of unlabeled data for labeling. There are three typical pool-based strategies as follows:
\par
\begin{itemize}
	\item \textbf{Random Select (RS) Method:} RS method can be viewed as a passive selection method, where unlabeled data are selected randomly at each iteration without any active query criterion. The RS method is often served as the baseline to be compared with other active selection methods.\par
	\item \textbf{Least Confidence (LC) Method \cite{zheng2002active}:} LC criterion selects the samples which are the least confident based on the posterior probabilities for all the classes.\par
	The well-known Max Entropy \cite{berger1996maximum,zhu2008active} is a popular uncertainty measurement widely used in previous studies of active learning: 
	$
	H(\mathbf{x})=-\sum_{y\in Y}P(y|\mathbf{x})\log P(y|\mathbf{x}),
	$
	where $P(y|\mathbf{x})$ is  a posteriori probability of $y$ given $\mathbf{x}$. We denote the output class as$y\in Y=\{y_1, y_2, ..., y_k\}$. $H(\cdot)$ is the uncertainty measurement function based on the entropy estimation of the classifier’s posterior distribution. The higher $H(\mathbf{x})$ is, the harder the example $\mathbf{x}$ seems to the classifier.
	\item \textbf{Margin based methods:} Margin based learning algorithm \cite{scheffer2001active, small2010margin} is an active learning algorithm which selects a hypothesis by minimizing a loss function using the margin of instance.
	For multi-class classification \cite{har2003constraint}, a widely accepted definition for multi-class margin is $p_{multiclass}(\mathbf{x}, y, f) = f_y(\mathbf{x}) - f_{\dot y}(\mathbf{x})$, where $y$ represents the true label and $\dot y = \arg\max_{y'\in \mathcal{Y}}\ f_{y'} (\mathbf{x})$ corresponds to the highest activation value such that $ \dot y \ne y$. So the samples we select can be formulated as 
	$$
		\mathbf{x}^*  = \arg\min_\mathbf{x}[f_{\hat y}(\mathbf{x})-f_{\tilde{y}}(\mathbf{x})],
	$$
	where $\hat y = \arg\max_{y'\in \mathcal{Y}}f_{y'}(\mathbf{x})$ represents the predicted label and $\tilde y = \arg\max_{y'\in \mathcal{Y}\backslash \hat y}f_{y'}(\mathbf{x})$ represents the label corresponding to the second highest activation value.
\end{itemize}

\section{Methodology}
In this section, we first present a passive learning framework which replaces the Jacobian-based method with other white-box attack approaches. Then, we introduce active learning into this framework to reduce the number of queries.
\subsection{A Passive Learning Framework}
The transfer-based framework \cite{papernot2017practical} firstly collects a very small set $S_0$ of inputs which are representatives of the input domain. Then it designs a network architecture $F$, which will be trained as the substitute model. The adversary applies a data augmentation technique on the current training set $S_\rho$ to produce a larger training set $S_{\rho+1}$ with more synthetic training points. The purpose of data augmentation is to learn about the decision boundary which is hidden in the deep neural networks. Our passive learning framework is summarized in Algorithm 1. The key operation is Step 5, in which we use certain method to craft samples for query. 
\begin{algorithm}[htbp]
\label{al1}
\caption{Substitute DNN Training}
        \textbf{INPUT:} target oracle $\tilde{O}$, 
a maximum number $\rho_{max}$ of training epochs, and an initial training set $S_0$.\\
        \textbf{OUTPUT:} a trained
substitute model $F$.
 
\begin{algorithmic}[1]

 \STATE Define architecture $F$;
 \FOR{$\rho=0; \rho<\rho_{max}; \rho++$}

 	\STATE $D\leftarrow D \cup \{(\mathbf{x},\tilde{O}(\mathbf{x}))|\mathbf{x}\in S_{add}\}$;

 	\STATE train $F$ with $D$;
 	\STATE craft $S_{add}$;
 	\STATE $S_{\rho +1} \leftarrow S_{\rho} \cup  S_{add}$;

 \ENDFOR
 \end{algorithmic}
\end{algorithm}\par

We denote $\tilde{O}(\mathbf{x})$ as the output of the target oracle $\tilde{O}$ queried with instance $\mathbf{x}$. The original method for crafting samples is the Jacobian-based method which constructs a sample along the direction of the gradient of the current substitute model and gradually learns the shape of the oracle's decision boundary. However, samples crafted by recent white-box attack methods contain more information about the decision boundary than those crafted by Jacobian-based augmentation. Instead of simply taking a fixed step along the direction of the gradient, these methods can adopt adaptive steps to make the constructed samples cross the decision boundary or directly solve an optimization problem to generate samples. So we use more advanced white-box attack methods to implement Step 5 of Algorithm 1, stated below.
\begin{itemize} 
	\item FGSM~\cite{goodfellow2014explaining}: $S_{add} = \{\mathbf{x}+\lambda \cdot \sign(\nabla_{\mathbf{x}}F(\mathbf{x}))|\mathbf{x} \in S_\rho\},$ where $\lambda$ is the hyper-parameter.
	\item Iterative Gradient Sign (IGS) \cite{kurakin2016adversarial}: 
 $$S_{add} = \{\mathbf{x}+ clip_\epsilon (\alpha \cdot \sign(\nabla_{\mathbf{x}}F(\mathbf{x}))) \},$$ where $clip_\epsilon(\cdot)$ performs a per-dimension clipping to constrain the result in the $\ell_{\infty}$ $\epsilon$-neighbourhood of the input $\mathbf{x}$.
	\item Fast Gradient Value (FGV) \cite{rozsa2016adversarial}:
$$S_{add} = \{\mathbf{x}+\lambda \cdot \nabla_{\mathbf{x}}F(\mathbf{x})|\mathbf{x} \in S_\rho\}$$
	\item JSMA~\cite{papernot2016limitations}:
Here we construct adversarial examples by modifying a limited number of pixels of the input image within the constrain of $\ell_2$-norm. 
	\item Deepfool~\cite{moosavi2016deepfool}:\par$$S_{add} = \Big\{\mathbf{x}+\frac{|F_l|\times |\nabla_{\mathbf{x}}F_l(\mathbf{x})|}{\|\nabla_{\mathbf{x}}F_l(\mathbf{x})\|_2^2} \times \sign(\nabla_{\mathbf{x}}F_l(\mathbf{x})) |\mathbf{x} \in S_\rho\Big\},$$ where $F_l$ is the $l$-th dimension of $F(\cdot)$.

	\item C\&W~\cite{carlini2017towards}:
For each instance $\mathbf{x}$ in $S_\rho$, we solve an optimization problem:\par
	$$  \quad\min \quad  \|\pmb{\delta}\|_2+c \cdot g(\mathbf{x}+\pmb{\delta})
	$$
	$$
	  \st \quad  \mathbf{x}+\pmb{\delta} \in[0,1]^n,\\
	$$
where $g(\mathbf{x}) = \max(\max_{i\ne l}(Z(\mathbf{x})_i)-Z(\mathbf{x})_t,-\kappa)$, $Z(\cdot)$ presents the softmax function and $\kappa$ is a constant to control the confidence. In our framework, we choose $\ell_2$-norm to constrain the size of the perturbation. The solutions to these optimization problems constitute $S_{add}$.\par

\end{itemize}\par
All these methods propose to generate adversarial samples across the decision boundary of the substitute model so that each round of black-box attack can accurately correct the model's parameters to make it more similar to the target oracle.

\subsection{Active Learning Strategy}
The above framework first trains a substitute DNN, then generates adversarial examples from the substitute DNN. However, it needs to query the oracle too many times, which is not allowed in real applications. So, we propose to use active learning to reduce the number of queries. The new procedure is summarized in Algorithm 2.\par
\begin{algorithm}[htbp]
\caption{Substitute DNN training using active learning strategy}
\label{al}
        \textbf{INPUT:} target oracle $\tilde{O}$, 
a maximum number $\rho_{max}$ of training epochs, and an initial training set $S_0$.\\
        \textbf{OUTPUT:} a trained
substitute model $F$.
 
\begin{algorithmic}[1]
 \STATE Define architecture $F$;
 \FOR{$\rho=0; \rho<\rho_{max}; \rho++$}
 
 	\IF{$\rho = 0$}
 	\STATE $D\leftarrow \{(\mathbf{x},\tilde{O}(\mathbf{x}))|\mathbf{x}\in S_\rho\}$;
 	\ELSE
 		\STATE $D_{add}\leftarrow \{(\mathbf{x},\tilde{O}(\mathbf{x}))|\mathbf{x}\in S_{add}\}$;
 		\STATE $D\leftarrow[D,D_{add}]$;
 	
 	\ENDIF
 	\STATE train $F$ with $D$;
 	\STATE craft $S_{add}$;
 	\STATE Use Active Learning strategy to generate a new $S_{add}$;
    \STATE $S_{\rho +1} \leftarrow S_{\rho} \cup  S_{add}$;

 \ENDFOR
 \end{algorithmic}
\end{algorithm}\par
In Step 10, $S_{add}$ can be crafted by any method mentioned above. The major difference is that in Step 11, we  use active learning strategy to select the most informative samples in the $S_{add}$. The motivation of introducing active learning is that different samples contribute differently to the learning process, i.e., if we add samples that can be classified by the current model with high confidence to the training set, then the decision boundary changes little. So, we use active learning strategy to select samples that help determine the decision boundary. This paper applies Random Select, Max Entropy and Margin based methods. \par
\begin{itemize}
	\item Random Select method (RS): We randomly select $k$ samples out of the initial set as the new $S_{add}$, where $k$ is the number of queries in each iteration.
	\item Max Entropy method (ME): We calculate 
 $H(\mathbf{x}) = -\sum_{y \in \mathcal{Y}}F(\mathbf{x})\log {F(\mathbf{x})}$ for each instance $\mathbf{x}$ in $S_{\rho}$, then we select $k$ samples with largest entropy as $S_{add}$.
 	\item Margin based method (MB): For each instance $\mathbf{x}$ in $S_{add}$, we denote the first and the second highest values of $F(\mathbf{x})$ by $h_1$ and $h_2$, and then set 
    $Dis_\mathbf{x} = h_1-h_2$. We take $k$ samples with smallest $Dis_\mathbf{x}$ as $S_{add}$.
\end{itemize}
  
  Random Select method is considered as the baseline method. Max Entropy method uses information entropy as a measure of the amount of information contained in a sample. 
 From a geometric point of view, this method gives priority to samples near the boundary, e.g., the confidence of samples far from the decision boundary is so high that we do not need to query the target oracle for their labels. Meanwhile, Margin based method achieves a similar benefit. $Dis_\mathbf{x}$ indicates the confidence of the substitute DNN about the unlabeled instance. The lower confidence it shows, the harder this instance seems to the current model. \par

 Although active learning strategy can select a small set of informative samples, it may introduce bias into the training set. For example, all the selected samples may concentrate in a small region of the input space. To address this limitation, we propose to increase the diversity of samples in each round of active learning. In this way, we are able to select samples that are informative and evenly distributed. We represent the diversity of a sample $\mathbf{x}$ by the distance between $\mathbf{x}$ and $S_\rho$, i.e., $ \min_{\mathbf{x}' \in S_\rho} \|\mathbf{x}-\mathbf{x}'\|$. We can integrate this criterion with all the active learning strategies above by considering the ranking of the instance selected by active learning methods and the ranking of the diversity simultaneously.\par

  Among the active learning methods mentioned above, Random Select method treats all the samples in $S_{add}$ as equal, so the ranking of each sample is the same. On the other hand, Max Entropy method and Margin based method calculate a score for each sample which implies a ranking. We combine the ranking of active learning strategies with that of the diversity to select a sample set for querying: 
\begin{itemize}
	\item RS + diversity: rank each instance $\mathbf{x}$ in $S_{add}$ according to $\min_{\mathbf{x}' \in S_\rho} \|\mathbf{x}-\mathbf{x}'\|$, and select the top $k$ largest samples.
	\item ME + diversity: rank each instance $\mathbf{x}$ according to $\Big\{ R\Big(-\sum_{y \in \mathcal{Y}}F(\mathbf{x})\log {F(\mathbf{x})}\Big)+R\Big(\min_{\mathbf{x}' \in S_\rho} \|\mathbf{x}-\mathbf{x}'\|\Big)\Big\}$, and select the top $k$ smallest samples.
 	\item MB + diversity: rank each instance $\mathbf{x}$ according to $\Big\{ r\Big(Dis_\mathbf{x}\Big)+R\Big(\min_{\mathbf{x}' \in S_\rho} \|\mathbf{x}-\mathbf{x}'\|\Big)\Big\}$, and select the top $k$ smallest samples.
 
\end{itemize}\par
	Here, we use $R$ to denote the ranking sorted from large to small, and $r$ to denote the ranking sorted from small to large.\par

\begin{table*}[!htb]
\centering

\label{table-1}
 \scalebox{1}[1]{
\begin{tabular}{|c|c|c|c|c|c|c|c|c|c|c|c|c|c|}
\hline
\multicolumn{2}{|c|}{} & \multicolumn{2}{c|}{FGSM} & \multicolumn{2}{c|}{IGS} & \multicolumn{2}{c|}{FGV} & \multicolumn{2}{c|}{Deepfool} & \multicolumn{2}{c|}{C\&W} & \multicolumn{2}{c|}{JSMA} \\ \hline
itr       & query      & Acc         & Simi         & Acc                   & Simi                  & Acc                   & Simi                   & Acc           & Simi           & Acc         & Simi            & Acc    & Simi   \\ \hline
0         &     100         &     0.4528       &     0.4521         &       0.4417               &         0.4764              &0.4577                      &0.4253                        &0.4259              &0.4009                &0.4533            &0.4197                        &0.4532       &0.4107        \\ \hline
1         &     200         &     0.3401       &     0.5628         &       0.2895               &         0.5684              &0.3483                      &0.5853                        &0.3257              &0.4894                &0.3792            &0.3577                        &0.2974       &0.6173        \\ \hline
2         &     400         &     0.2085       &     0.7521         &       0.2642               &         0.7648              &0.2412                      &0.7451                        &0.2161              &0.7415                &0.2373            &0.7504                        &0.2384       &0.7936        \\ \hline
3         &     800         &     0.2201       &     0.7706         &       0.2061               &         0.7684              &0.1989                      &0.7701                        &0.1753              &0.7679                &0.2156            &0.7564                        &0.1936       &0.8362        \\ \hline
4         &     1600        &    0.2253        &    0.7865          &       0.2427               &         0.7962              &0.1783                      &0.8136                        &0.1242              &0.8628                &0.1688            &0.8328                        &0.1873       &0.8635        \\ \hline
5         &     3200        &    0.1439        &    0.8330          &       0.1426               &         0.8286              &0.1209                      &0.8460                        &0.0832              &0.8969                &0.1196            &0.8572                        &0.1639       &0.8935        \\ \hline
6         &     6400        &    0.1289        &    0.8623          &       0.1317               &         0.8530              &0.1374                      &0.8595                        &0.0801              &0.8739                &0.0693            &0.9166                        &0.1373       &0.9126        \\ \hline
7         &     12800       &   0.0906         &   0.8682           &       0.0810               &         0.8778              &0.1326                      &0.8877                        &0.0639              &0.9283                &0.0617            &0.9310                & 0.1299      & 0.9263  \\ \hline
8         &     25600       &   0.0652         &   0.8821           &       0.0656               &         0.8940              &0.0752                      &0.9053                        &   0.0625           &    \textbf{0.9419}            &     \textbf{0.0563}       &   0.9217                &   0.1108    &  0.9183      \\ \hline

\end{tabular}}
\caption{Results of using FGSM, Iterative Gradient Sign (IGS), JSMA, Fast Gradient Value (FGV), Deepfool and C\&W to craft samples for querying on MNIST. The evaluation metrics are Acc and Simi. The results on this table are averaged over 10 runs.}
\end{table*}

\begin{table}[!htb]
\centering

\label{table-2}
 \scalebox{1}[1]{
\begin{tabular}{|c|c|c|c|}
\hline
\# of Iterations      & 0      & 3      & 9      \\ \hline
FGSM     & 0.5173 & 0.3746 & 0.2362 \\ \hline
IGS      & 0.4762 & 0.3349 & 0.2169 \\ \hline
FGV      & 0.5238 & 0.3676 & 0.2681 \\ \hline
Deepfool & 0.5017 & 0.3574 & 0.2031 \\ \hline
C\&W     & 0.5362 & 0.2717 & \textbf{0.1939} \\ \hline
JSMA     & 0.4846 & 0.2907 & 0.2301 \\ \hline
\end{tabular}
}
\caption{Results on the Fashion-MNIST dataset. The evaluation metric is Acc. The results on this table are averaged over 10 runs.}
\end{table}

\begin{table*}[!htb]
\centering

\label{table-3}
 \scalebox{1}[1]{
\begin{tabular}{|c|c|c|c|c|c|c|c|c|c|c|c|c|c|}
\hline
\multicolumn{2}{|c|}{} & \multicolumn{2}{c|}{FGSM} & \multicolumn{2}{c|}{IGS} & \multicolumn{2}{c|}{FGV} & \multicolumn{2}{c|}{Deepfool} & \multicolumn{2}{c|}{C\&W} & \multicolumn{2}{c|}{JSMA} \\ \hline
itr       & query      & Acc         & Simi         & Acc                   & Simi                  & Acc                   & Simi                   & Acc           & Simi           & Acc         & Simi            & Acc    & Simi   \\ \hline
0         &     100         &     0.5641       &     0.4712         &       0.5361               &         0.5351              &0.4714                      &0.4736                        &0.4533              &0.5361                &0.5713            &0.4613                        &0.4893       &0.5136        \\ \hline
1         &     200         &     0.5315       &     0.4616         &       0.5215               &         0.5271              &0.4461                      &0.4361                        &0.4174              &0.5713                &0.5135            &0.4719                        &0.4616       &0.5574        \\ \hline
2         &     400         &     0.4712       &     0.5135         &       0.4612               &         0.5713              &0.4533                      &0.6345                        &0.3671              &0.6471                &0.5105            &0.5531                        &0.4941       &0.6184        \\ \hline
3         &     800         &     0.4513       &     0.7174         &       0.4471               &         0.5582              &0.4168                      &0.6643                        &0.3577              &0.7473                &0.4723            &0.6341                        &0.4536       &0.7145        \\ \hline
4         &     1600        &    0.4462        &    0.7436          &       0.4463               &         0.6364              &0.3757                      &0.6851                        &0.3416              &0.7747                &0.4613            &0.6234                        &0.4531       &0.7747        \\ \hline
5         &     3200        &    0.4164        &    0.7481          &       0.4136               &         0.7485              &0.3983                      &0.7275                        &0.3174              &0.7431                &0.4123            &0.6572                        &0.4212       &0.7557        \\ \hline
6         &     6400        &    0.3963        &    0.7764          &       0.3857               &         0.7796              &0.3557                      &0.7758                        &0.2747              &0.7512                &0.3681            &0.7156                        &0.3873       &0.7683        \\ \hline
7         &     12800       &   0.3713         &   0.8361           &       0.3641               &         0.7979              &0.3164                      &0.7843                        &0.2685              &0.7736                &0.3512            &0.7351                & 0.3791      & 0.7791  \\ \hline
8         &     25600       &   0.3413         &   0.8412           &       0.3813               &         0.7837              &0.3264                      &0.7951                        &\textbf{0.2579}              &0.7971                &0.3377            &0.7257                &   0.3681    &  \textbf{0.7975}      \\ \hline

\end{tabular}}
\caption{Results of using FGSM, Iterative Gradient Sign (IGS), JSMA, Fast Gradient Value (FGV), Deepfool and C\&W to craft samples for querying on CIFAR-10. The evaluation metrics are Acc and Simi. The results on this table are averaged over 5 runs.}
\end{table*}

\begin{figure*}[!htb] 

\subfigure[Acc vs. Number of queries]{
\includegraphics[width=6cm]{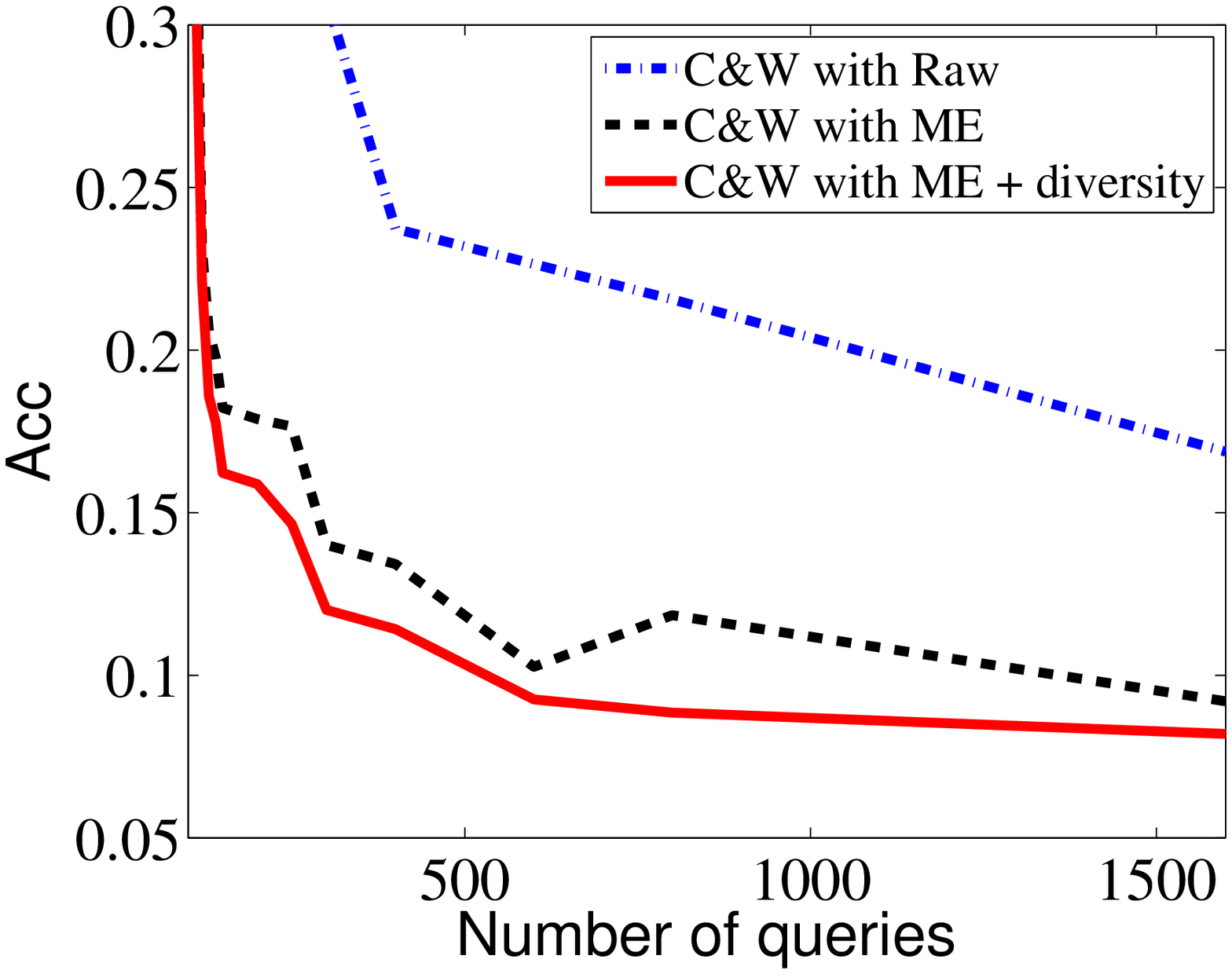}\\
\includegraphics[width=6cm]{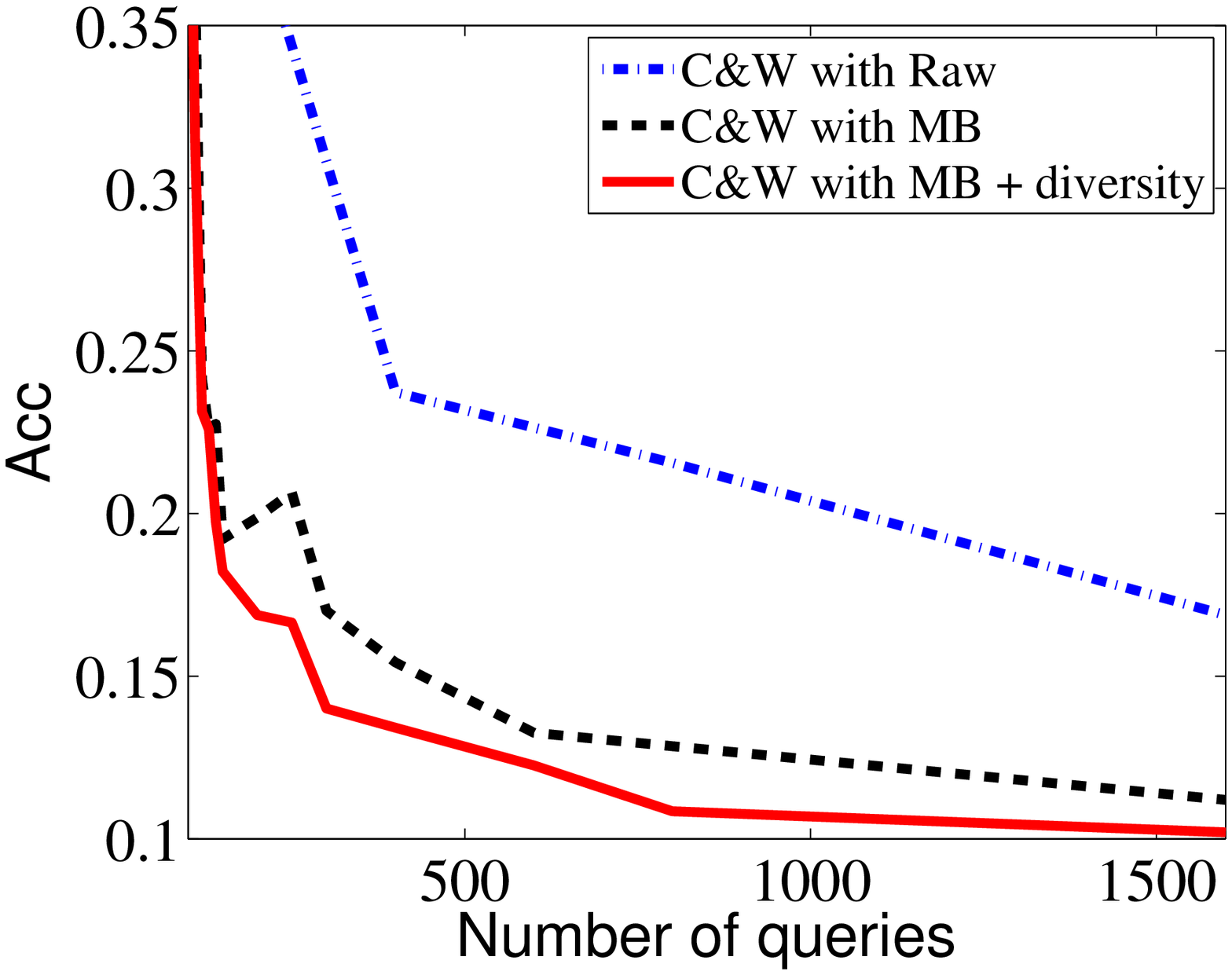}\\
\includegraphics[width=6cm]{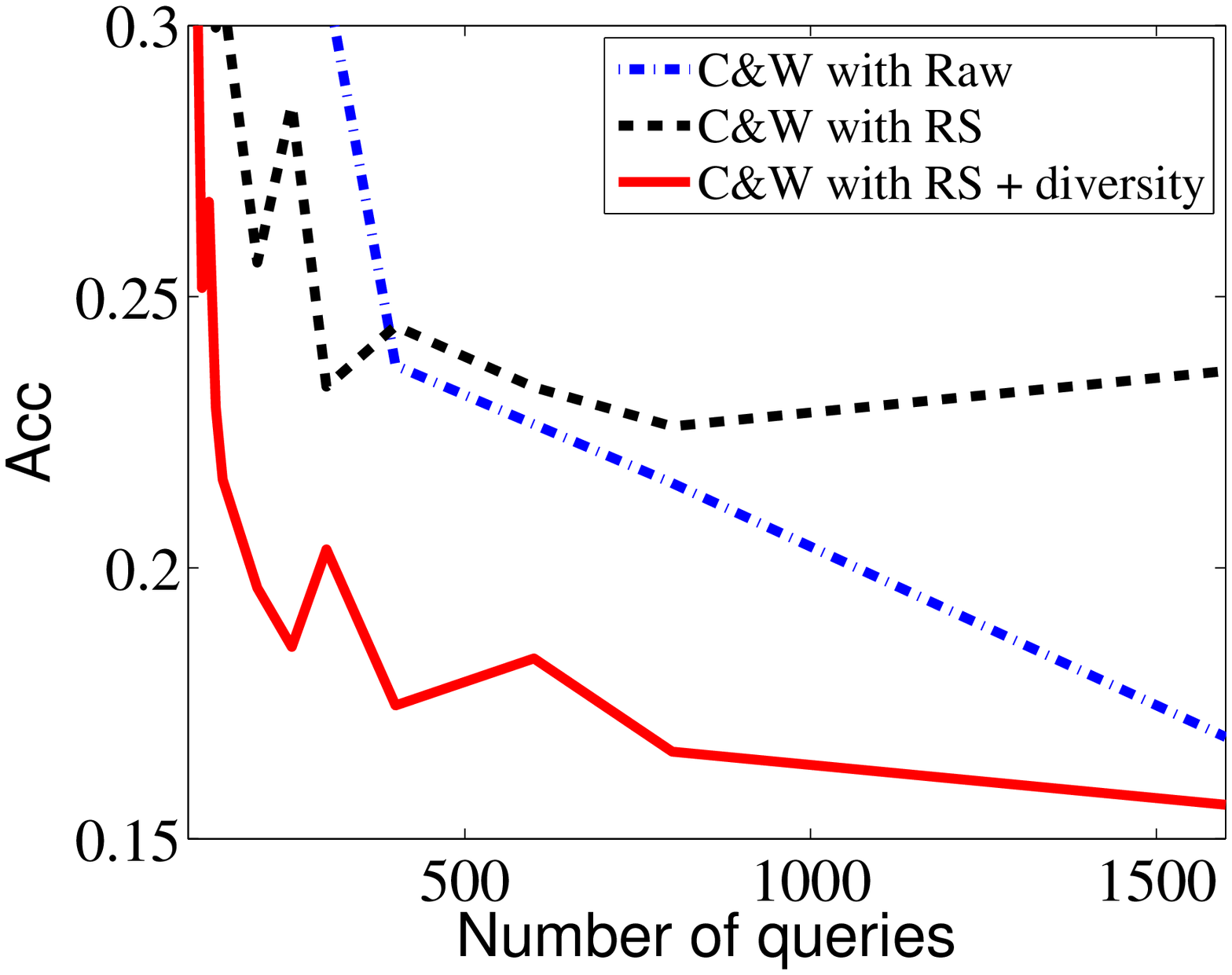}\\
}
\subfigure[Simi vs. Number of queries]{
\includegraphics[width=6cm]{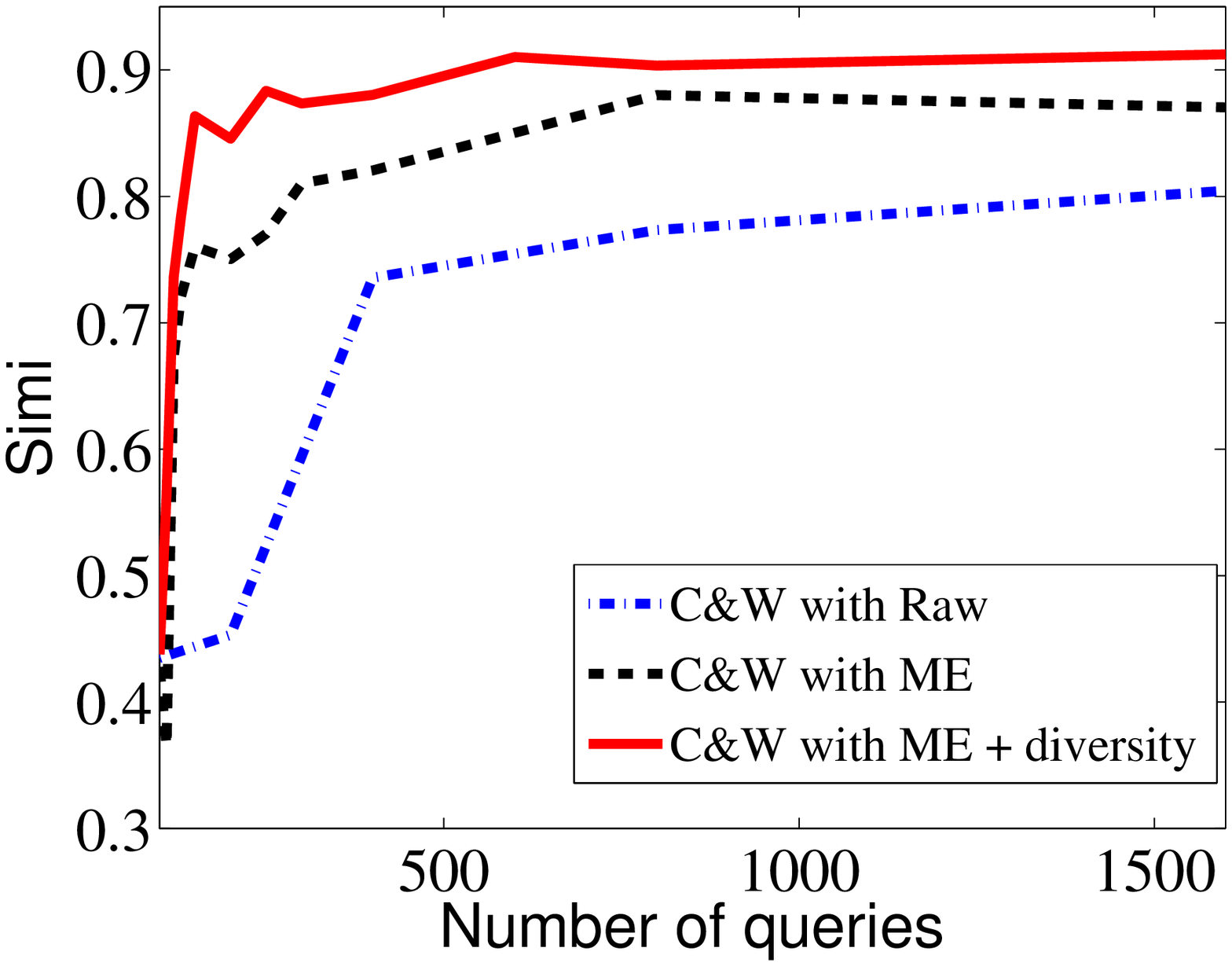}\\
\includegraphics[width=6cm]{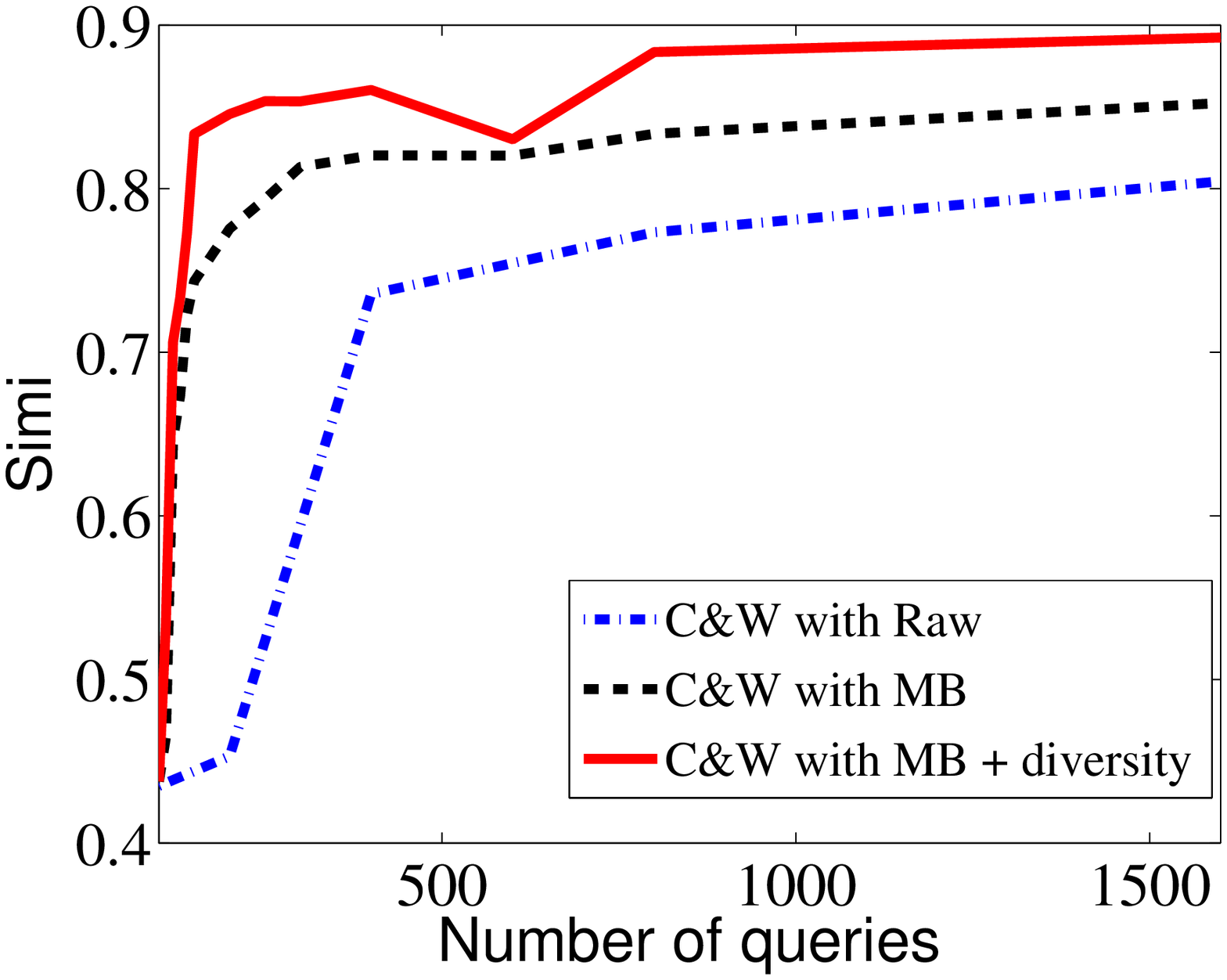}\\
\includegraphics[width=6cm]{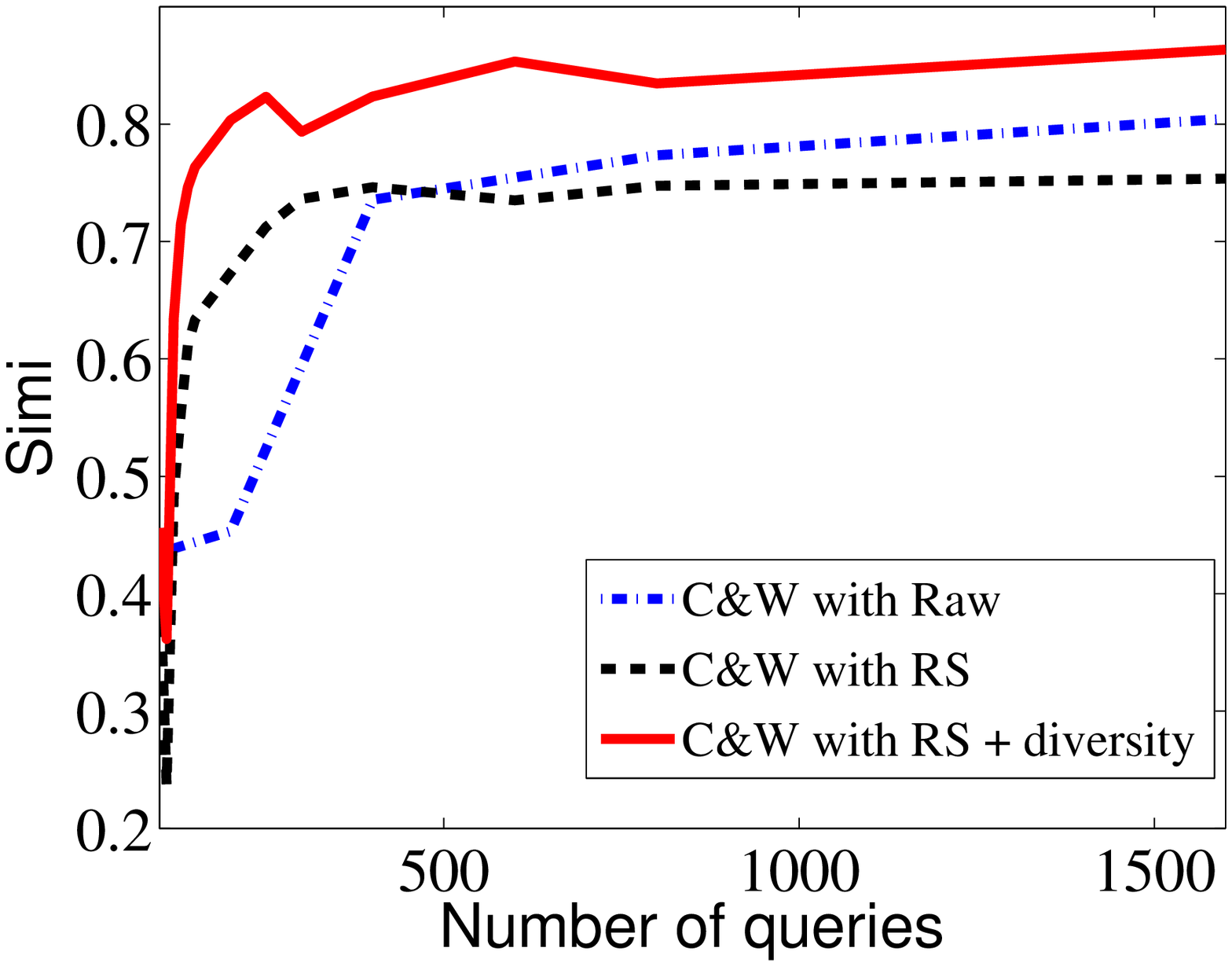}\\
}

\caption{Black-box attack by combining C\&W and active learning strategy on MNIST (ME: Max Entropy method, MB: Margin based method, RS: Random Select method, diversity: our proposed active learning strategy). The curves on this Fig are averaged over 10 runs.}
\label{fig:2} 
\end{figure*} 

\begin{figure*}[!htb] 

\subfigure[Acc vs. Number of queries]{
\includegraphics[width=6cm]{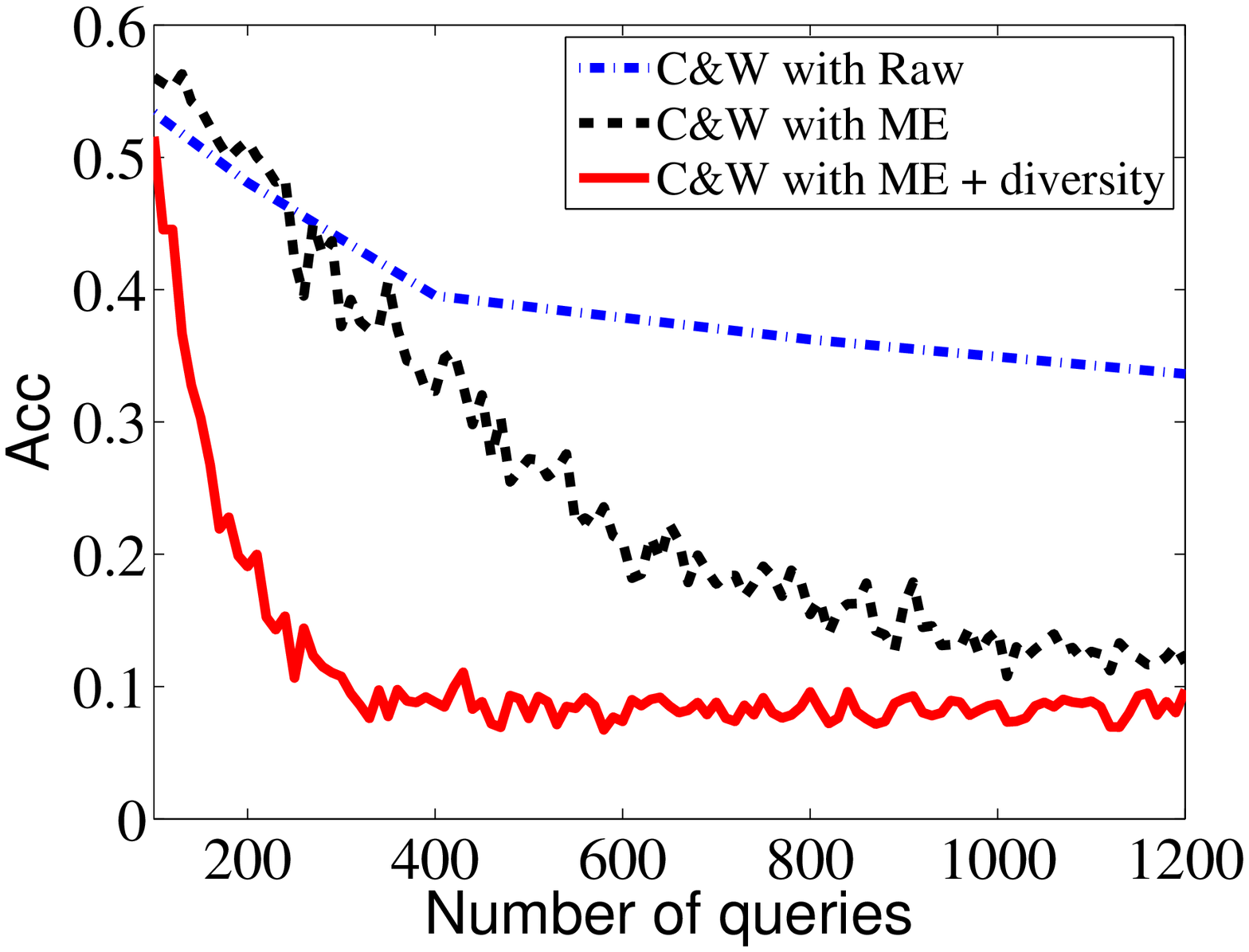}\\
\includegraphics[width=6cm]{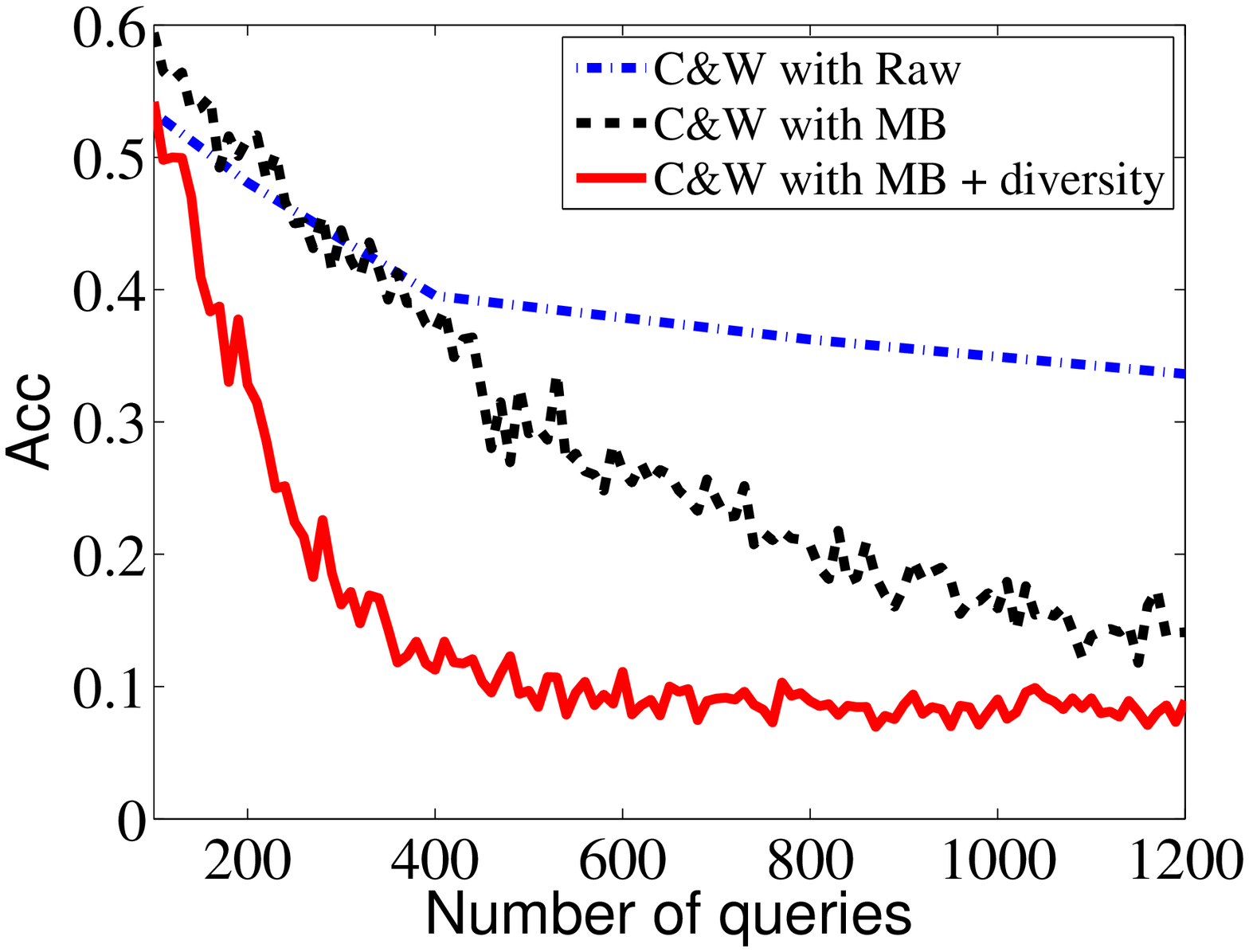}\\
\includegraphics[width=6cm]{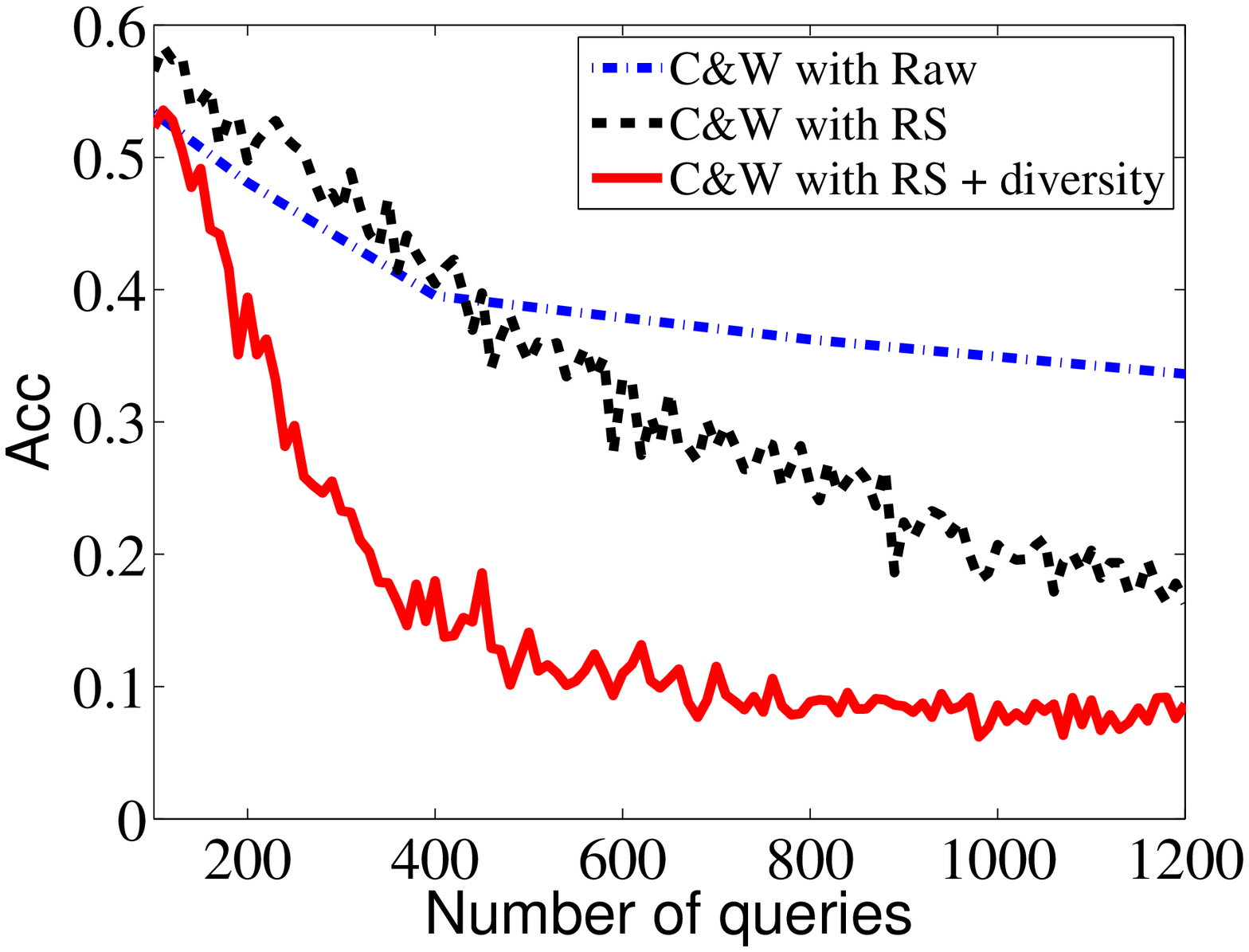}\\
}
\subfigure[Simi vs. Number of queries]{
\includegraphics[width=6cm]{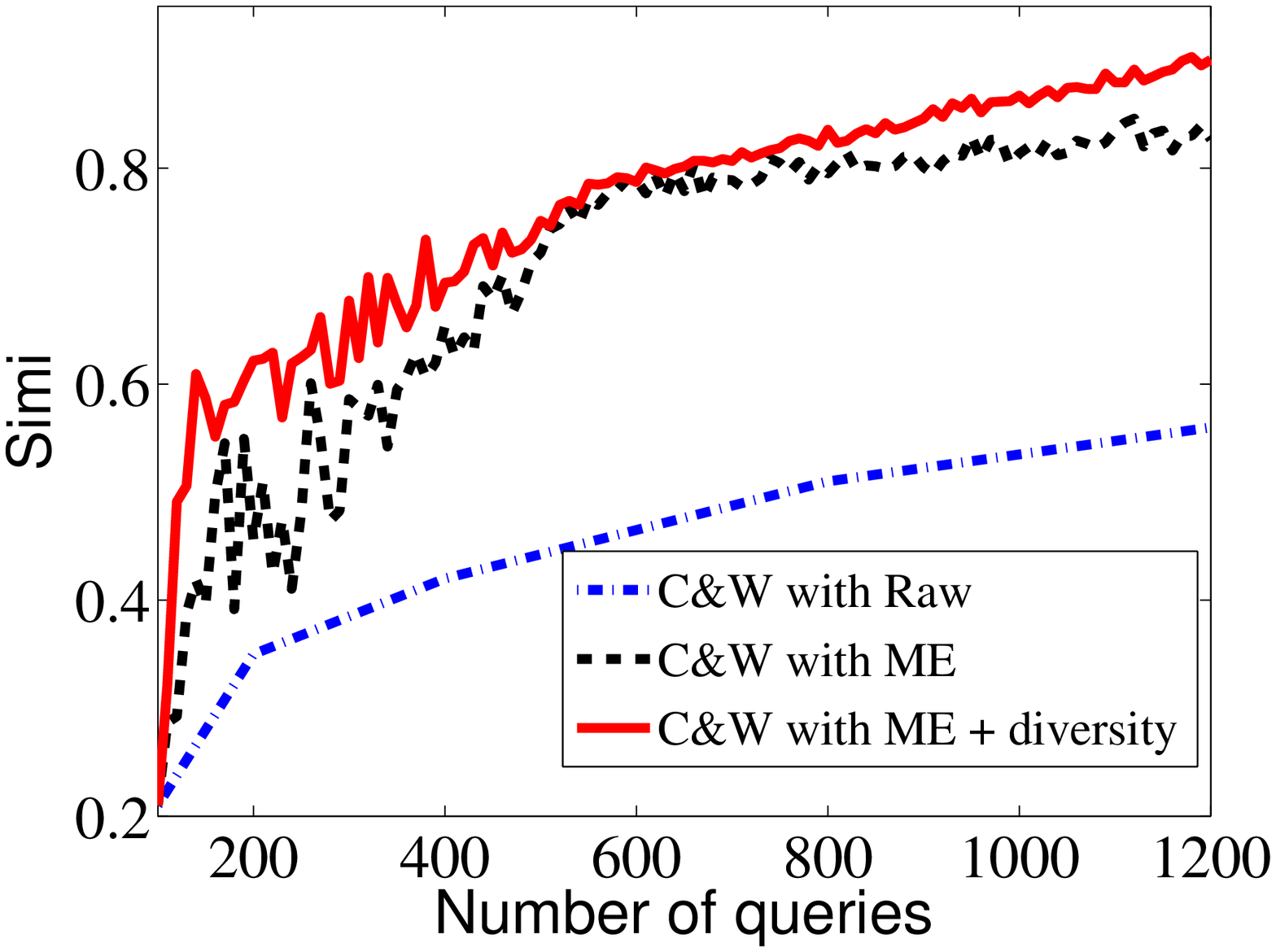}\\
\includegraphics[width=6cm]{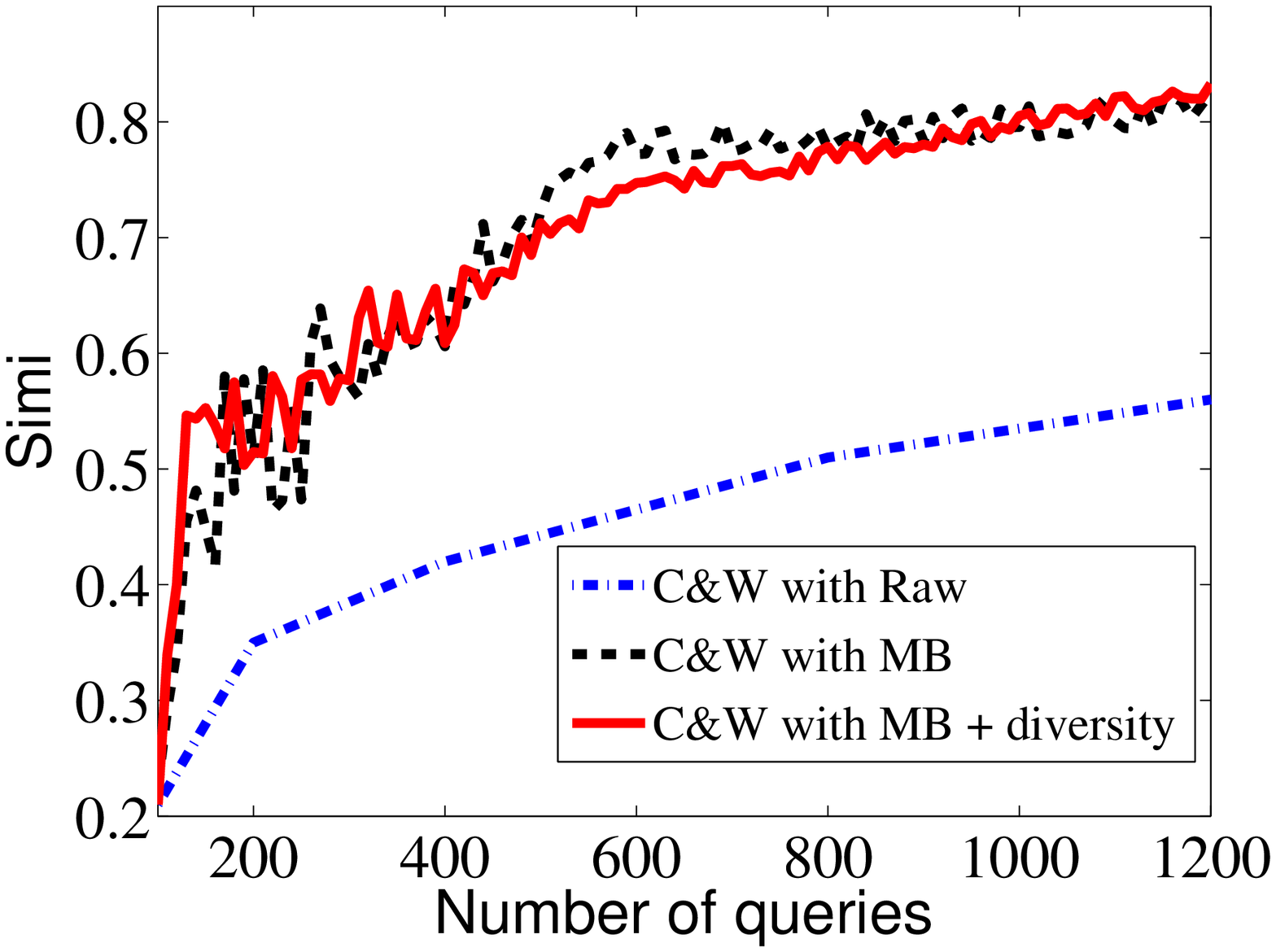}\\
\includegraphics[width=6cm]{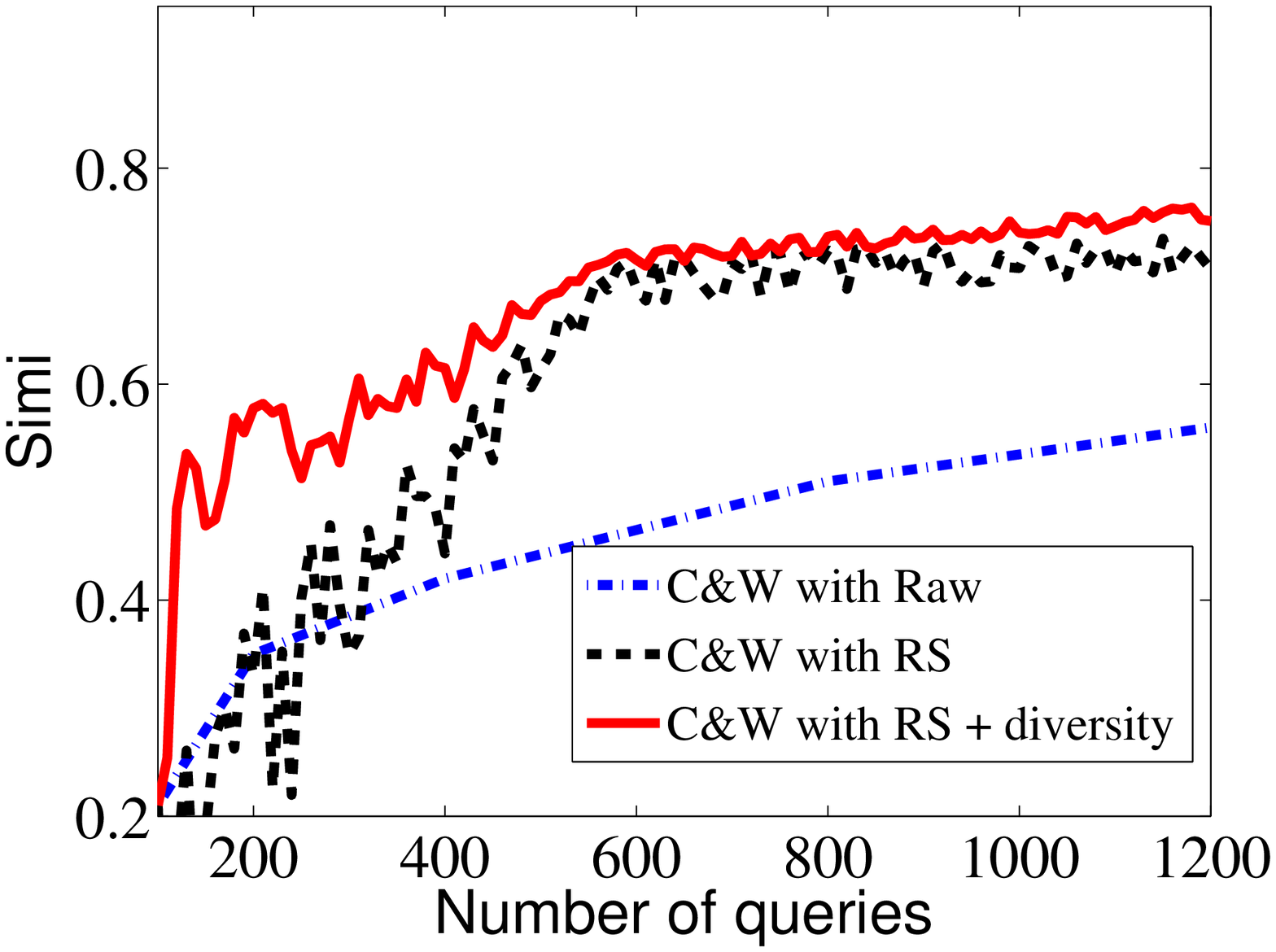}\\
}
\caption{Black-box attack by combining C\&W and active learning strategy on CIFAR-10 (ME: Max Entropy method, MB: Margin based method, RS: Random Select method, diversity: our proposed active learning strategy). The curves on this Fig are averaged over 5 runs.}
\label{fig:2} 
\end{figure*}

\begin{figure*}[!htb] 

\subfigure[Acc vs. Number of queries]{
\begin{minipage}[b]{0.5\textwidth}
\centering
\includegraphics[width=7.9cm]{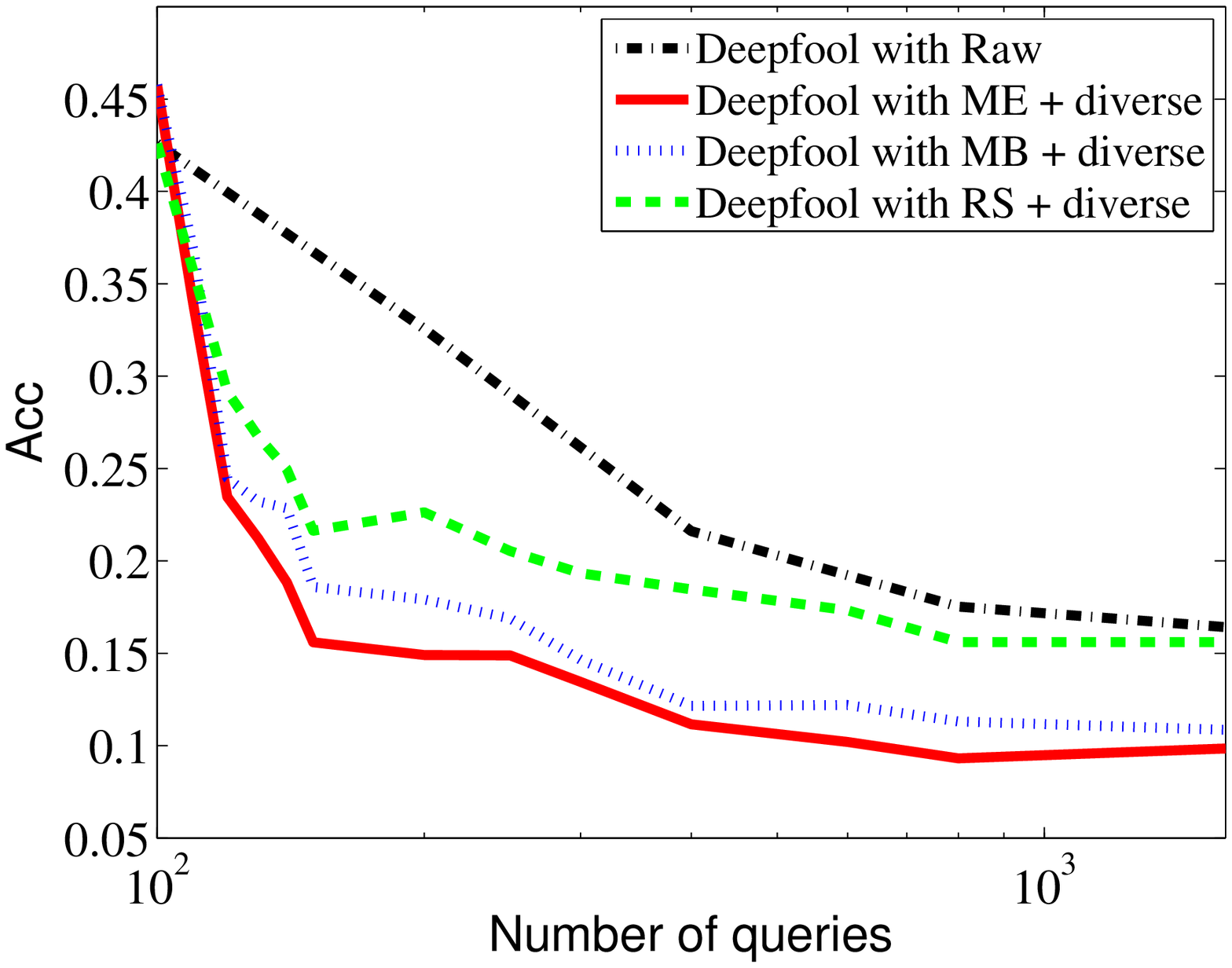}
\end{minipage}
}
\subfigure[Simi vs. Number of queries]{
\begin{minipage}[b]{0.5\textwidth}
\centering
\includegraphics[width=7.9cm]{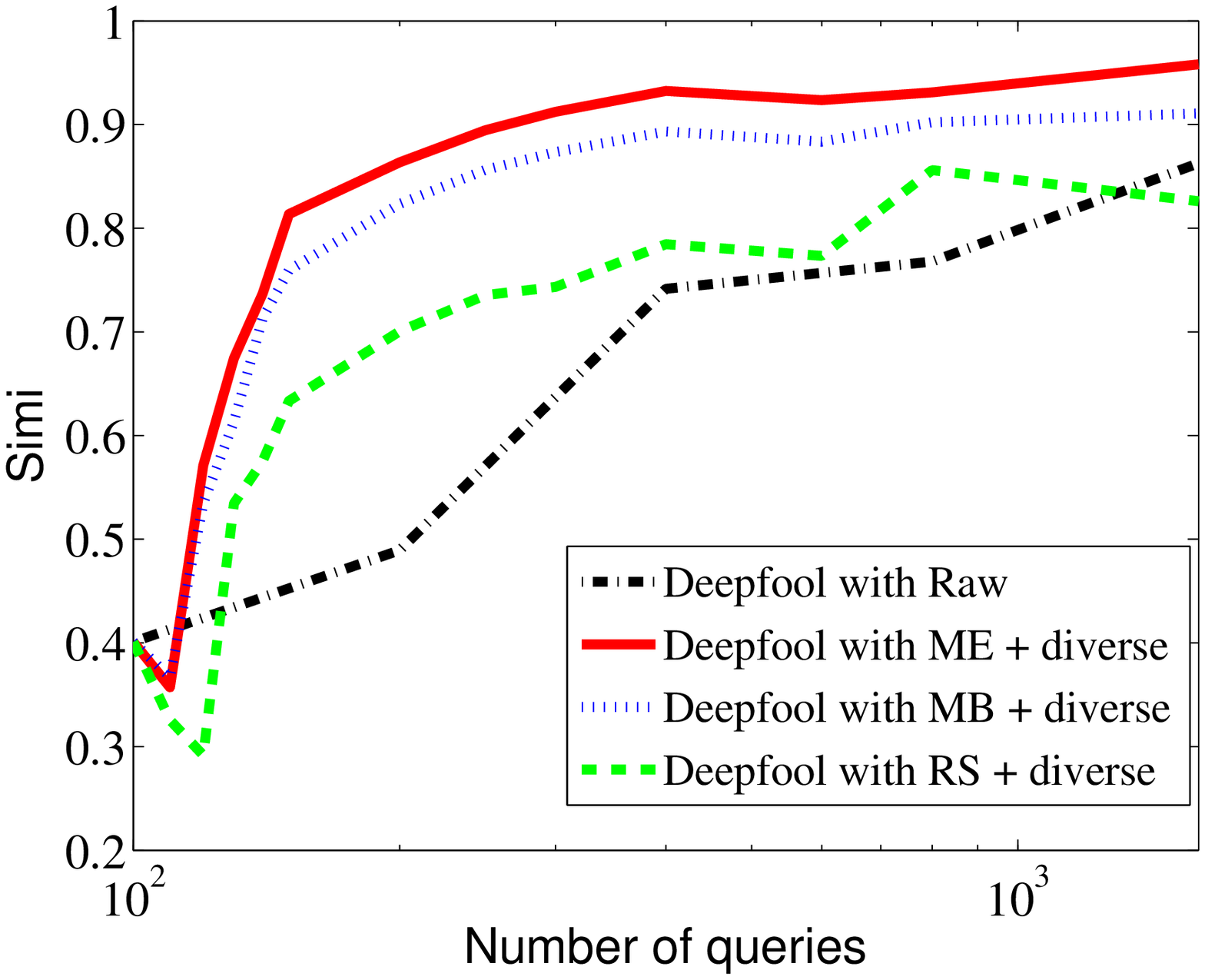}
\end{minipage}
}
\caption{Black-box attack by combining Deepfool and active learning strategy on MNIST (ME: Max Entropy method, MB: Margin based method, RS: Random Select method, diversity: our proposed active learning strategy). The curves on this Fig are averaged over 10 runs.}
\label{fig:1} 
\end{figure*} 
\begin{figure*}[!htb] 

\subfigure[Acc vs. Number of queries]{
\begin{minipage}[b]{0.5\textwidth}
\centering
\includegraphics[width=7.9cm]{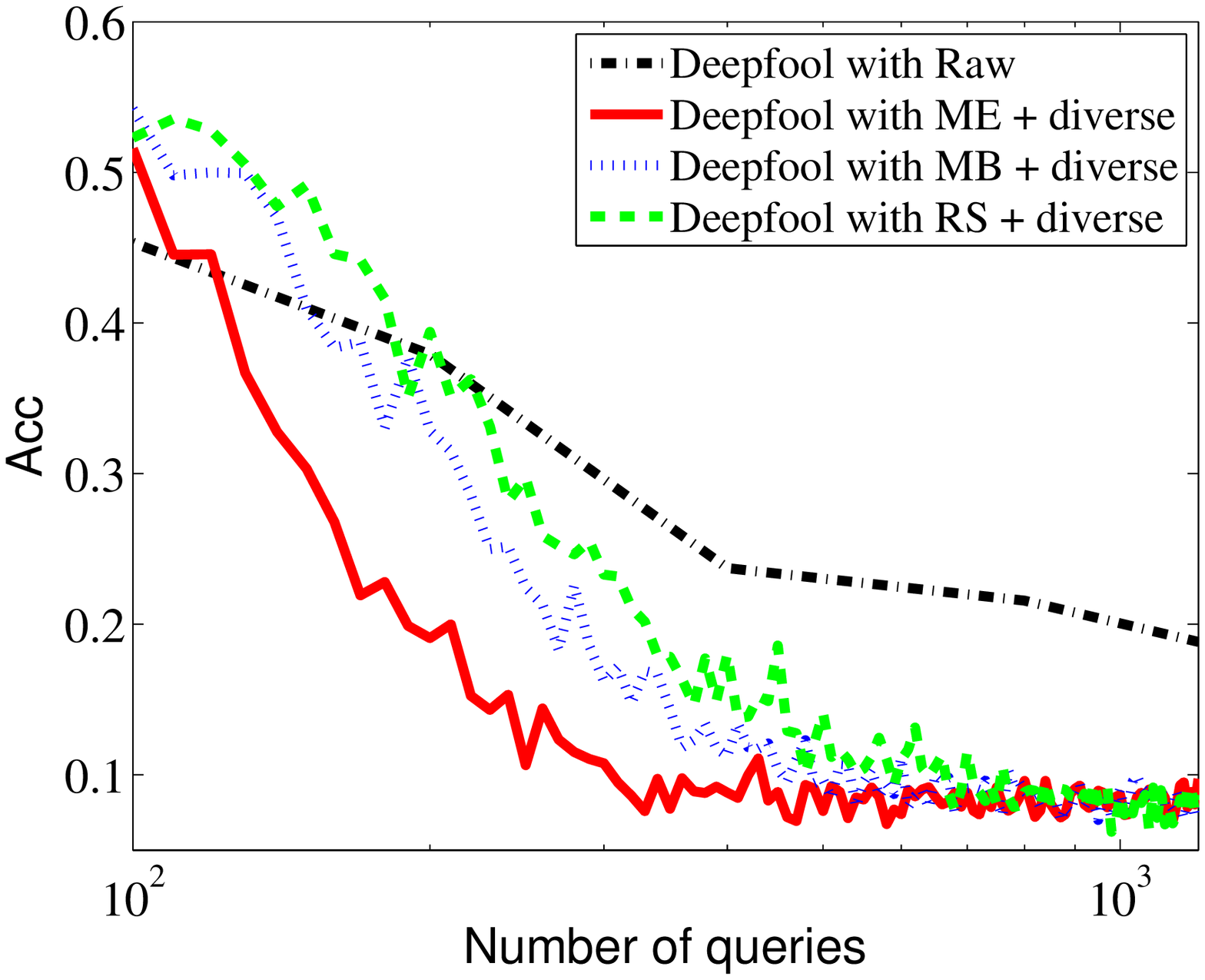}
\end{minipage}
}
\subfigure[Simi vs. Number of queries]{
\begin{minipage}[b]{0.5\textwidth}
\centering
\includegraphics[width=7.9cm]{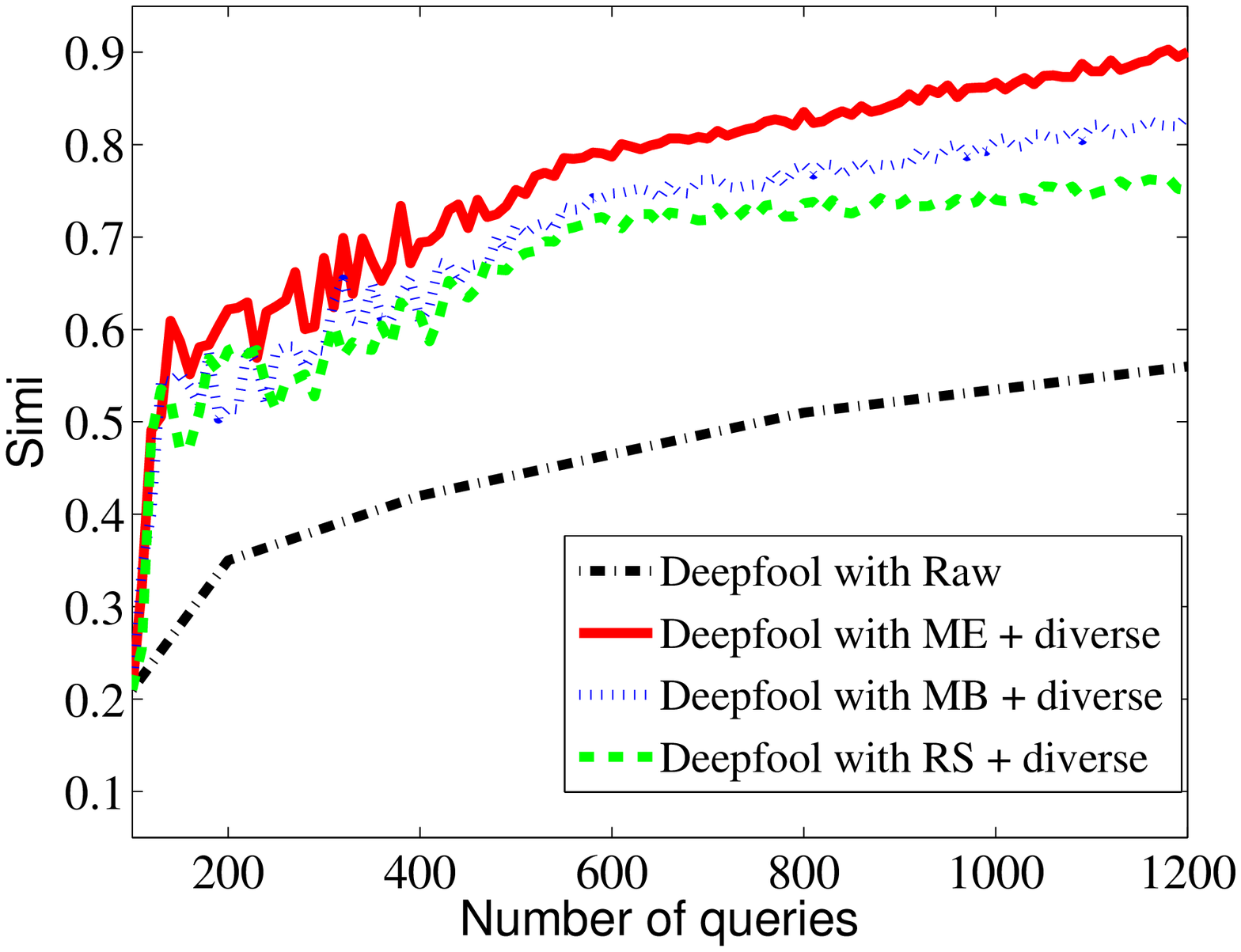}
\end{minipage}
}

\caption{Black-box attack by combining Deepfool and active learning strategy on CIFAR-10 (ME: Max Entropy method, MB: Margin based method, RS: Random Select method, diversity: our proposed active learning strategy). The curves on this Fig are averaged over 5 runs.}
\label{fig:2} 
\end{figure*} 

\section{Experiment}

In this section, we first compare the performance of different data augmentation methods within the passive learning framework. Then, we validate the performance of our algorithm which considers active learning and diversity of the query set simultaneously.
\subsection{Setup}
	We evaluate the performance of our algorithm on three datasets, i.e., MNIST \cite{lecun1998mnist}, Fashion-MNIST \cite{xiao2017fashion} and CIFAR-10 \cite{krizhevsky2009learning}. We random select 100 samples as the initial training set (with 10 samples from each class) for each dataset. We assume adversaries can collect such a limited sample set from the oracle task. 
    
    The metric we used to evaluate the attack pattern is divided into two parts: Accuracy of the target oracle (Acc) and Similarity (Simi). Acc is an indicator of the success rate of attacks (the lower, the better). We use FGSM to generate adversarial examples over the substitute model and denote the accuracy of oracle when tested with these adversarial examples as Acc. Simi represents the similarity between our substitute model and the oracle (the higher, the better). We query the oracle with the entire dataset and treat the result as a new dataset. We denote the accuracy of the substitute model when tested with this dataset by Simi. \par

	For dataset MNIST and Fashion-MNIST, we use a pre-trained CNN model with accuracy $99.24\%$ as the target oracle and a simple CNN model which contains a convolutional layer of 32 convolution kernels, a max-pooling layer, and a fully-connected layer as the substitute network. The structure of the target oracle is invisible to our algorithm. The parameters of white-box attack methods are set according to the suggestion in their original papers. The parameter $\lambda$ of FGSM attack which aims to generate query samples is set to be 0.2. The parameters $\epsilon$ and $\alpha$ of Iterative Gradient Sign are set to be $0.2$ and $10$. The $\lambda$ of Fast Gradient Value is set to be 0.2. The iteration of C\&W is set as 100 and the learning rate is set as 0.005. The goal of these parameters is to constrain the adversarial samples not far from the original samples. For dataset CIFAR-10, we use a pre-trained Resnet \cite{he2016deep} model whose accuracy is more than $91\%$ as the target oracle and train a substitute VGG-19 \cite{simonyan2014very} network to attack the oracle. The parameters of the white-box attack methods are set to be the same as that used in MNIST. 
    
\subsection{Passive Learning Framework}
	We construct the samples with different white-box attack methods and compare the performance of the substitute models trained thereby.  
     In black-box attack, the number of queries is the main cost. The more we query, the more likely to cause the target oracle's attention. The experimental results on the MNIST dataset are shown in Table 1, from which we observe that when using C\&W or Deepfool method to craft query samples, the iteration we need for an effective attack is less than FGSM and other methods. We can also verify that with the same number of queries, the substitute model trained by data generated with C\&W or Deepfool method has a higher attack success rate and a higher similarity with the oracle. Results on the Fashion-MNIST dataset are consistent, and due to the similarity between MNIST and Fashion-MNIST,  we only report partial results in Table 2. We also present a full result over CIFAR-10 in Table 3. As shown in Table 3, the performance of C\&W and Deepfool is better than the rest of the attack methods.\par

	The reason for the better performance of C\&W and Deepfool is that these methods solve an optimization problem to craft samples which can cross the boundary of the current model. In contrast, other methods like FGSM construct samples from the original samples along the direction of the gradient towards the decision boundary but may not cross the boundary. This confirms our thought that there is more information about the boundary within the samples crafted by solving optimization problems than samples crafted by gradient based attack methods.

\subsection{Active Learning Strategy}
Since C\&W and Deepfool show a better performance than other methods in previous experiments, we combine these two methods with different active learning strategies in this part.\par

Fig. 2 shows the performance of the active learning method, as well as our improved version that takes the diversity of samples into consideration, where the parameter $k$ is set to be 10.  The Raw algorithm in Fig. 2 follows the setting in \cite{papernot2017practical} which doubles the number of queries in the first few iterations each time, and then uses reservoir sampling method \cite{vitter1985random} which is a Random Selection strategy in the later few iterations to make the number of queries grow linearly.

As we can see in Fig. 2(a), the Acc of each active learning method is lower than that of the Raw algorithm except RS method. The reason is that the Raw algorithm queries more samples than the RS methods in the first few iterations. However,  as the number of iterations increases, the effects of Raw algorithm and RS method become comparable. Furthermore, in all cases, Acc of our improved version is lower than that of the original active learning method. For example, to achieve $10\%$ Acc, the original Max Entropy algorithm queries over 1600 times, our improved version only queries $600$ times, while the Raw algorithm queries more than 10000 times.
The curves on Simi in Fig. 2(b) also validate our motivation that the diversity of sample set can effectively identify different parts of the decision boundary. We observe that Simi increases much more quickly when we take diversity into consideration. For example, to achieve $85\%$ Simi, the original Max Entropy algorithm queries over $600$ times, while the improved version only needs $200$ times. This situation suggests that, the distortions in the decision boundary of our substitute models are more similar to that of oracle and considering the diversity of query set can help modify the decision boundary better than only using active learning algorithm. Due to the similarity between MNIST and Fashion-MNIST, we omit the curves of Fashion-MNIST in this paper.

We also tested our algorithm on CIFAR-10 and presented the results of the active learning method, as well as our improved version that takes the diversity of samples into consideration. The Acc of our improved version is lower than that of the original active learning method. For example, to achieve $10\%$ Acc, the original Max Entropy algorithm queries over $5000$ times, our improved version only queries $400$ times, while the Raw algorithm queries more than $25600$ times. 

In summary, while the original active learning algorithm (Max Entropy method and Margin based method) can be used to reduce the number of queries significantly, our improved version can further boost the performance.
\par
In Fig. 4, we change the method from C\&W to Deepfool (Deepfool achieve as good performance as C\&W in the experiments of the previous passive learning framework) and report the performance of our improved algorithm which considers active learning strategy and diversity simultaneously. Again, the results show that the combination of active learning and the diversity of samples indeed reduces the number of queries in transfer-based black-box attack significantly. For example, when the number of queries reaches up to $400$, the Acc of Max Entropy algorithm is $10\%$ lower than that of the Raw algorithm, and the Simi is $20\%$ higher.

In Fig. 5, We also tested our algorithm on CIFAR-10. We use Deepfool to generate training data for the substitute model. The Max Entropy strategy shows the best performance. The accuracy of the target oracle against adversarial examples is lower than $10\%$ and the number of queries which the original algorithm need is more than $10000$, while our active learning strategy can reduce to less than $1000$.

\section{Conclusion and Future Work}
In this paper, we have tested a number of white-box attack methods and found that C\&W attack and Deepfool yield the overall best performance. In addition, we introduced active learning to address the query-efficiency issue occurred in transfer-based attack. To alleviate the bias caused by active learning, we propose to maximize the diversity of query set and our empirical study verifies its effectiveness. In the future, we will apply our method to a variety of machine learning models, rather than neural networks and apply more advanced active learning strategy.

\section*{ACKNOWLEDGMENT}
This work was partially supported by the National Key R\&D Program of China (2018YFB1004300), YESS (2017QNRC001), and the Collaborative Innovation Center of Novel Software Technology and Industrialization.

\bibliographystyle{IEEEtran}  
\bibliography{attack}  

\end{document}